\documentclass[10pt,twocolumn,letterpaper]{article}

\usepackage{cvpr}
\usepackage{times}
\usepackage{epsfig}
\usepackage{graphicx}
\usepackage{amsmath}
\usepackage{amssymb}
\usepackage{subfig}
\usepackage{multirow}
\usepackage[]{algorithm2e}
\usepackage{booktabs}

\usepackage{pifont}% http://ctan.org/pkg/pifont

\usepackage{xcolor}
\usepackage{color}
\definecolor{darkgreen}{rgb}{0,0.6,0.2}
\newcommand{\comment}[1]{{{#1}}}
\newcommand{\commentM}[1]{{{#1}}}
\newcommand{\commentMG}[1]{{#1}}

\newcommand{\ADGN}{DGP}

\usepackage{enumitem}
\newcommand\Tstrut{\rule{0pt}{2.6ex}}

\usepackage{authblk}
\makeatletter
\renewcommand\AB@affilsepx{, \protect\Affilfont}
\makeatother
\usepackage[pagebackref=true,breaklinks=true,letterpaper=true,colorlinks,bookmarks=false]{hyperref}

\cvprfinalcopy % *** Uncomment this line for the final submission

\def\cvprPaperID{5745} % *** Enter the CVPR Paper ID here

\ifcvprfinal\fi
\begin{document}

%%%%%%%%% TITLE
\title{Rethinking Knowledge Graph Propagation for Zero-Shot Learning}

\author[1]{Michael Kampffmeyer\thanks{Indicates equal contribution.}}
\author[2]{Yinbo Chen$^*$}
\author[3]{Xiaodan Liang\thanks{Corresponding Author.}}
\author[4]{Hao Wang}
\author[5]{Yujia Zhang}
\author[6]{Eric P. Xing}
\affil[1]{UiT The Arctic University of Norway}
\affil[2]{Tsinghua University}
%\makeatletter
%\renewcommand\AB@affilsepx{,\\\Affilfont}
%\makeatother
\affil[3]{Sun Yat-sen University}
%\makeatletter
%\renewcommand\AB@affilsepx{,\protect\Affilfont}
%\makeatother
\affil[4]{Massachusetts Institute of Technology}
\makeatletter
\renewcommand\AB@affilsepx{,\\\Affilfont}
\makeatother
\affil[5]{Institute of Automation, Chinese Academy of Sciences}
\affil[6]{Carnegie Mellon University}

\iffalse
\author{First Author\\
Institution1\\
Institution1 address\\
{\tt\small firstauthor@i1.org}
\and
Second Author\\
Institution2\\
First line of institution2 address\\
{\tt\small secondauthor@i2.org}
}\fi

\setlength{\textfloatsep}{8pt plus 0pt minus 3pt}
\maketitle
\thispagestyle{empty}

%%%%%%%%% ABSTRACT
\begin{abstract}
   Graph convolutional neural networks have recently shown great potential for the task of zero-shot learning. These models are highly sample efficient as related concepts in the graph structure share statistical strength allowing generalization to new classes when faced with a lack of data. 
   %However, we find that the extensive use of Laplacian smoothing at each layer in current approaches can easily dilute the knowledge from distant nodes and consequently decrease performance.
   \commentM{However, multi-layer architectures, which are required to propagate knowledge to distant nodes in the graph, dilute the knowledge by performing extensive Laplacian smoothing at each layer and thereby consequently decrease performance.}
   % in zero-shot learning.
   In order to still enjoy the benefit brought by the graph structure while preventing dilution of knowledge from distant nodes, we propose a Dense Graph Propagation (\ADGN) module with carefully designed direct links among distant nodes.
    {\ADGN} allows us to exploit the hierarchical graph structure of the knowledge graph through additional connections. These connections are added based on a node's relationship to its ancestors and descendants. A weighting scheme is further used to weigh their contribution depending on the distance to the node to improve information propagation in the graph. Combined with finetuning of the representations in a two-stage training approach our method outperforms state-of-the-art zero-shot learning approaches.
\end{abstract}

%%%%%%%%% BODY TEXT
\section{Introduction}
With the ever-growing supply of image data, from an ever-expanding number of classes, there is an increasing need to use prior knowledge to classify images from unseen classes into correct categories based on semantic relationships between seen and unseen classes. This task is called zero-shot image classification. %To obtain satisfactory performance on this task, it is 
Crucial to this task is precise modeling of class relationships based on prior class knowledge. Previously, prior knowledge has been incorporated in form of semantic descriptions of classes, such as attributes~\cite{akata2015evaluation,romera2015embarrassingly,long2017zero} or word embeddings~\cite{socher2013zero,frome2013devise,li2017zero}, or by using semantic relations such as knowledge graphs~\cite{palatucci2009zero,rohrbach2011evaluating,salakhutdinov2011learning,lu2016unsupervised}. Approaches that use knowledge graphs are less-explored and generally assume that unknown classes can exploit similarity to known classes.
Recently, the benefit of hybrid approaches that combine knowledge graph and semantic class descriptions has been illustrated~\cite{wang2018zero}.
%Unlike class attributes, knowledge graphs and word embeddings are readily available for large number of concepts. 

\begin{figure*}[htpb] 
\subfloat[Graph Propagation]{\includegraphics[width=0.44\linewidth,trim={4.4cm 2cm 3cm 0.5cm},clip]{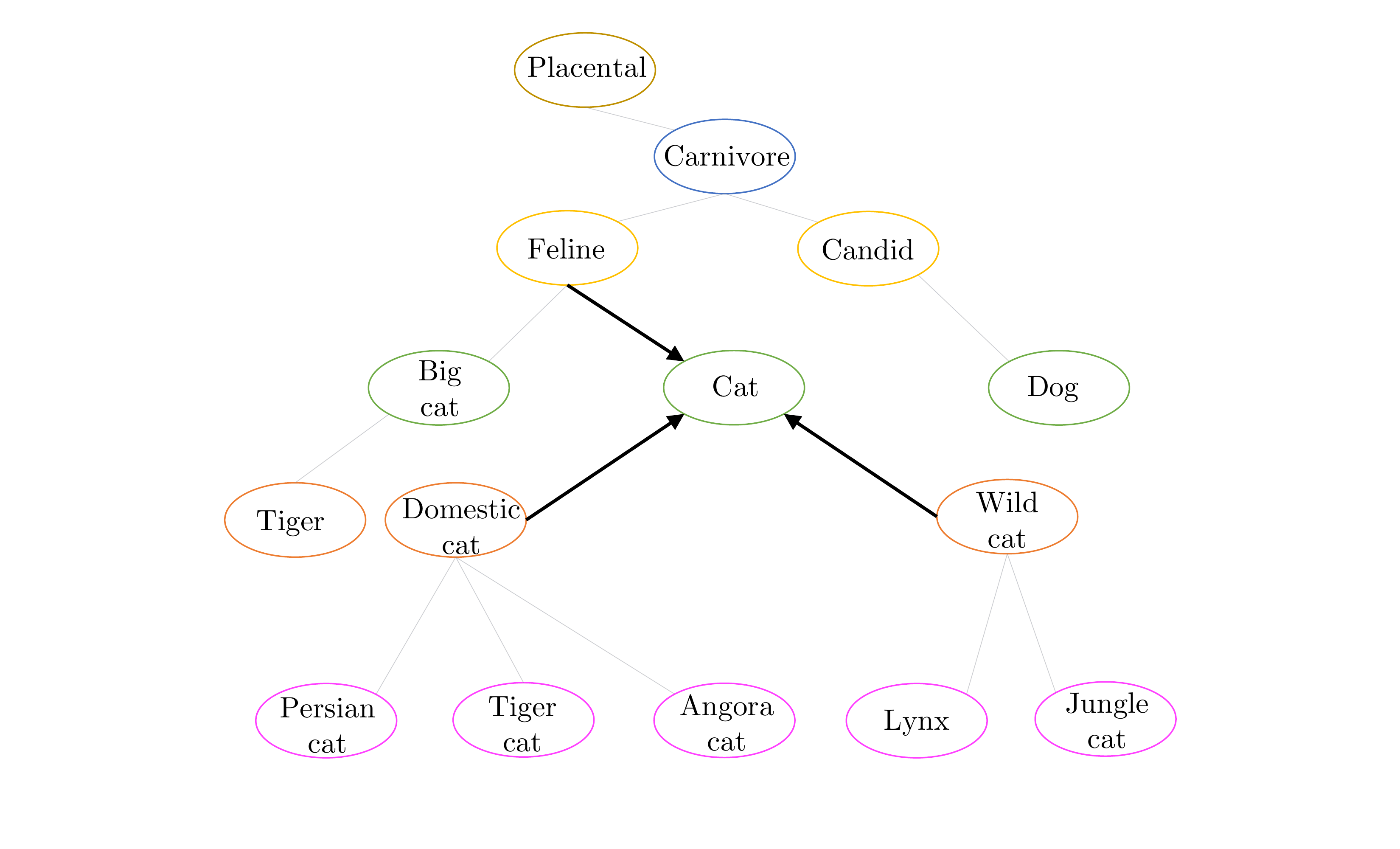}}
\subfloat[Dense Graph Propagation]{\includegraphics[width=0.50\linewidth,trim={2cm 2cm 3cm 0.5cm},clip]{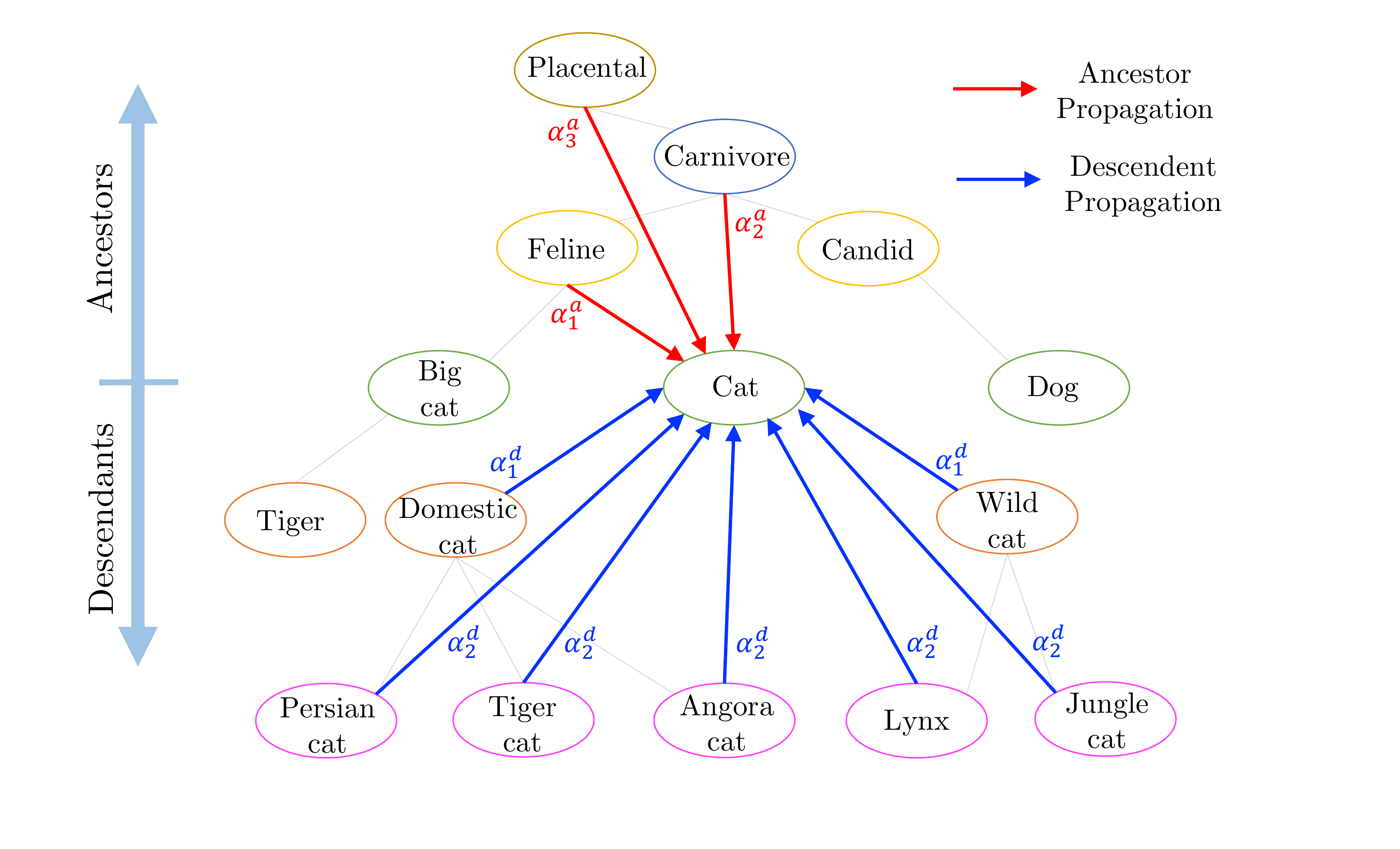}}
\caption{a) Illustration of graph propagation in a GCN~\cite{kipf2016semi} for node 'Cat'. Here, graph propagation represents the knowledge that a node receives in a single layer for previous approaches. b) Proposed dense graph propagation for node 'Cat'. The node receives knowledge from all its descendants during the descendant phase (blue arrows) and its ancestors during the ancestor phase (red arrows). This leads to a densely connected graph where knowledge can directly propagate between related nodes. The learned weights $\alpha_k^a$ and $\alpha_k^d$ are used to weigh nodes that are $k$-hops away from a given node in the ancestor and the descendants phase, respectively.
}\vspace{-1mm}
\label{fig:graph_connection}
\end{figure*}

The current state-of-the-art by Wang \etal~\cite{wang2018zero} processes the \commentM{unweighted} knowledge graph by exploiting recent developments in neural networks for non-Euclidean spaces, such as graph and manifold spaces~\cite{bronstein2017geometric}.
%approach by Wang et al.~\cite{wang2018zero} processes knowledge graphs by making use of recent developments in applying neural network techniques to non-euclidean spaces. 
%A deep graph convolutional neural network (GCN)~\cite{kipf2016semi} is used and the problem is phrased as weight regression, where the GCN is trained to regress classifier weights for each class. 
\commentM{A deep graph convolutional neural network (GCN)~\cite{kipf2016semi} is used and the problem is phrased as a regression task, where the GCN is trained to output a classifier for each class by regressing real-valued weight vectors. These weight vectors correspond to the last layer weights of a pretrained convolutional neural network (CNN) and can be viewed as logistic regression classifiers on top of the feature extraction produced by the CNN.}
GCNs balance model complexity and expressiveness with a simple scalable model relying on the idea of message passing, i.e. nodes pass knowledge to their neighbors. However, these models were originally designed for classification tasks, albeit semi-supervised, an arguably simpler task than regression. In recent work, it has been shown that GCNs perform a form of Laplacian smoothing, where feature representations will become more similar as depth increases leading to easier classification~\cite{li2018deeper}. In the regression setting, instead, the aim is to exchange information between nodes in the graph and extensive smoothing is not desired as it dilutes information and does not allow for accurate regression. For instance, in a connected graph all features in a GCN with $n$ layers will converge to the same representation as $n\to \infty$ under some conditions, hence washing out all information~\cite{li2018deeper}.

Therefore, we argue that this approach is not ideal for the task of zero-shot learning and that the number of layers in the GCN should be small in order to avoid smoothing. %Intuitively, since every node averages information, the information will become more and more diluted.
%To further illustrate our intuition, we can view this problem in analogy with the Telephone game, a popular children's game. In this game a group of players successfully have to pass a secret message from the first player in the group to the last player by whispering the message one player at a time, however, the message generally gets diluted by intermediate players.
%Similarly, intermediate nodes in the graph dilute knowledge through the smoothing operations, resulting in worse zero-shot performance. 
We illustrate this phenomenon in practice, by showing that a shallow GCN consistently outperforms previously reported results. %We employ a model-of-models framework by training the method to predict a set of logistic regression classifier for each class on top of a set of extracted features produced by a CNN.
Choosing a small number of layers, however, has the effect that knowledge will not propagate well through the graph. A $1$-layer GCN for instance only considers neighbors that are two hops away in the graph such that only immediate neighbors influence a given node. 
Thus, we propose a dense connectivity scheme, where nodes are connected directly to descendants/ancestors in order to include distant information. 
%These connections allow us to propagate information without many smoothing operations but, leads to the problem that all descendants/ancestors are weighed equally when computing the regression weight vector for a given class. However, intuitively, nodes closer to a given node should have higher importance. \commentM{Further, this would effectively remove the majority of the structure in the graph as all ancestors would be considered one-hop neighbors.} 
%\commentM{To allow the model to preserve the structure in the graph, we further propose a weighting scheme that considers the distance between nodes in order to weigh the contribution of different nodes.}
\commentM{These new connections allow us to propagate information without over-smoothing but remove important structural information in the graph since all descendants/ancestors would be included in the one-hop neighborhood and would be weighed equally when computing the regression weight vector for a given class. To address this issue, we further propose a weighting scheme that considers the distance between nodes in order to weigh the contribution of different nodes. This allows the model to not only recover the original structure in the graph but further provides an additional degree of flexibility that enhances the inference capabilities of our model.}
%To remedy this, we extend this framework by adding a weighting scheme that considers the distance between nodes in order to weigh the contribution of different nodes, \commentM{thereby enabling the model to preserve the structure in the graph.} 
Introducing distance-based shared weights also has the benefit that it only adds a minimal amount of parameters, is computationally efficient, and balances model flexibility and restrictiveness to allow good predictions for the nodes of the unseen classes.  
Fig.~\ref{fig:graph_connection} illustrates the difference in the way knowledge is propagated in this proposed Dense Graph Propagation ({\ADGN}) module compared to a GCN layer.
%\commentY{Maybe we can just say it like ``we further proposed the weighing scheme for DGP, which has the benefits that...'', but not ``We have A, to remedy something, we added B to it.''?}

To allow the feature extraction stage of the pre-trained CNN to adjust to the newly learned classifiers we propose a two-phase training scheme. In the first step, the {\ADGN} is trained to predict the last layer CNN weights. In the second phase, we replace the last layer weights of the CNN with the weights predicted by the {\ADGN}, freeze the weights and finetune the remaining weights of the CNN by optimizing the cross entropy classification loss on the seen classes. 

%We further consider the problem of domain-shift in zero-shot learning, which is the problem that current methods often struggle to perform well on both seen and unseen classes~\citep{chao2016empirical}. Seen and unseen classes can be considered overlapping domains with a set of shared appearances, however, they also exhibit domain differences~\citep{kodirov2017semantic}. To remedy this and allow the feature extraction stage of the pre-trained CNN to adjust to the newly learned classifiers we propose a two-phase training scheme. In the first step, the {\ADGN} is trained to predict the last layer CNN weights. In the second phase, we replace the last layer weights of the CNN with the weights predicted by the {\ADGN}, freeze the weights and fine-tune the remaining weights of the CNN by optimizing the cross entropy classification loss on the seen classes. 
%to predict the {\ADGN} output in order to allow the CNN feature representation to adjust to the predicted weights \commentY{This is not what fine-tuning does. Fine-tuing optimizes the cross entropy loss of classification while substituting predicted weights for original weights}, incorporating the implicit constraint of the knowledge graph. \commentY{For this paragraph, fine-tuning sounds like an approach for solving domain-shift, but actually it should be more like a training trick (or not even trick, but natural thing), and is a 2-stage version of joint-training which is common in few-shot tasks. And I am still quite confused about why fine-tuning can exactly solve domain difference.}

Our main contributions are the following:
\begin{itemize}[noitemsep]
    \vspace{-2mm}
    \item An analysis of our intuitions for zero-shot learning and an illustration of how these intuitions can be combined to design a {\ADGN} that outperforms previous state-of-the-art approaches.\footnote{The source code for the experiments performed in this paper is available at: \url{https://github.com/cyvius96/adgpm}.}
    %\url{https://github.com/cyvius96/adgpm}. \commentM{Upload to some anonymous dropbox folder and add link?}}.
    \item Our {\ADGN} module, which explicitly exploits the hierarchical structure of the knowledge graph to perform zero-shot learning by efficiently propagating knowledge through the proposed dense connectivity structure.
    \item \commentM{A novel weighting scheme for {\ADGN} where weights are learned based on the distance between nodes.}
    \item Experimental results on various splits of the 21K ImageNet dataset, a popular large-scale dataset for zero-shot learning. We obtain relative improvements of more than $50\%$ over previously reported best results.
\end{itemize}

%We present an analyzes of our intuitions for zero-shot learning and illustrate how these intuitions can be combined to design a {\ADGN} that outperforms previous zero-shot learning results.
%To summarize, the key benefit of the proposed {\ADGN} module is that it explicitly exploits the hierarchical structure of the knowledge graph in order to perform zero-shot learning by more efficiently propagating knowledge through the proposed dense connectivity structure.
%Specifically, we perform experiments on various splits of the 21K ImageNet dataset, a large scale dataset that is commonly used for zero-shot learning. 
%On the 21K classes of ImageNet, we obtain relative improvements of more than $50\%$ over the previously reported best results, and our proposed {\ADGN} improves on the 1-layer GCN by $7.1\%$.
%In a realistic setting zero-shot models should generalize well to the unseen classes but still perform well on seen classes~\citep{norouzi2013zero}. We therefore also evaluate the ability of the models to perform well on the seen classes and illustrate that the performance on the seen classes benefits from our finetuning approach.

%%%%%%%%%%%%%%%%%%%%%%%%%%%%%%%%%%%%%%%%%%%%%%%%
%%%%%%%%%%%%%%%% Related Work %%%%%%%%%%%%%%%%%%
%%%%%%%%%%%%%%%%%%%%%%%%%%%%%%%%%%%%%%%%%%%%%%%%
\section{Related Work}

\textbf{Graph convolutional networks} are a class of graph neural networks, based on local graph operators~\cite{bruna2013spectral,defferrard2016convolutional,kipf2016semi}. Their advantage is that their graph structure allows the sharing of statistical strength between classes making these methods highly sample efficient.
After being introduced in Bruna \etal~\cite{bruna2013spectral}, they were extended with an efficient filtering approach based on recurrent Chebyshev polynomials, reducing their computational complexity to the equivalent of the commonly used CNNs in image processing operating on regular grids~\cite{defferrard2016convolutional}. Kipf \etal~\cite{kipf2016semi} further proposed simplifications to improve scalability and robustness and applied their approach to semi-supervised learning on graphs. \commentM{Their approach is termed graph convolutional network (GCN) and provides the foundation for the model in this paper.}

\textbf{Zero-shot learning} has in recent years been considered from various set of viewpoints such as manifold alignment~\cite{deutsch2017zero,li2017zero}, linear auto-encoder~\cite{kodirov2017semantic}, and low-rank embedded dictionary learning approaches~\cite{ding2017low}, using semantic relationships based on attributes~\cite{misra2017red,socher2013zero,frome2013devise} and relations in knowledge graphs~\cite{wang2018zero,mensink2012metric,rohrbach2011evaluating,palatucci2009zero}.
One of the early works~\cite{larochelle2008zero} proposed a method based on the idea of a model-of-models approach, where a model is trained to predict class models based on their description.
Each class is modeled as a function of its description. 
This idea has recently been used in another work in Wang \etal~\cite{wang2018zero}, the work most similar to our own, where a graph convolutional neural network is trained to predict logistic regression classifiers on top of pre-trained CNN features in order to predict unseen classes. Their approach has yielded impressive performance on a set of zero-shot learning tasks and can, to the author's knowledge, be considered the current state-of-the-art.
%\commentM{Link to meta learning?} 

%%%%%%%%%%%%%%%%%%%%%%%%%%%%%%%%%%%%%%%%%%%%%%%%
%%%%%%%%%%%%%%%%%% Approach %%%%%%%%%%%%%%%%%%%%
%%%%%%%%%%%%%%%%%%%%%%%%%%%%%%%%%%%%%%%%%%%%%%%%
\section{Approach}
Here we first formalize the problem of zero-shot learning and provide information on how a GCN model can be utilized for the task. We then describe our proposed model {\ADGN}. 
%\commentY{I was thinking if we should make our model independent of GCN series, since the only common things is averaging (which is in fact sym-norm vs. ave-norm)}

\begin{figure*}
  \centering
  \includegraphics[trim={0.5cm 8.9cm 3.6cm 0cm},clip, width=1\linewidth]{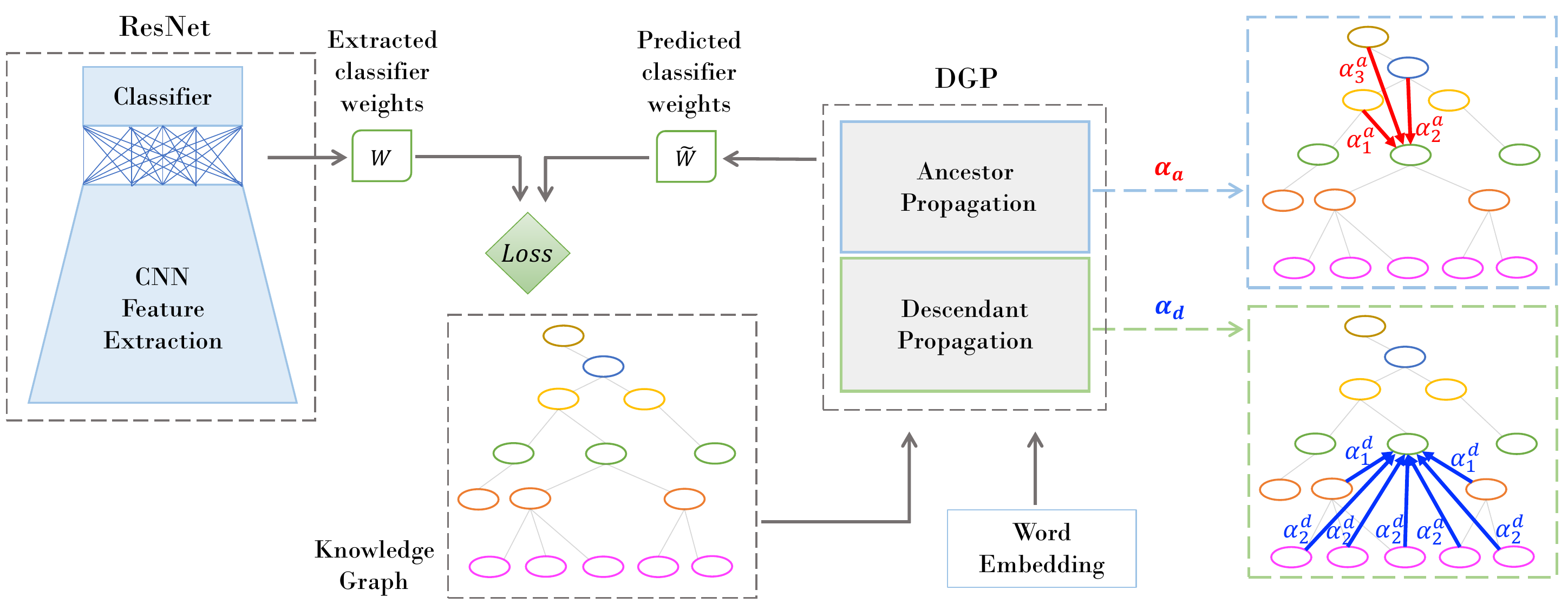}
  \caption{{\ADGN} is trained to predict classifier weights $W$ for each node/class in a graph. The weights for the training classes are extracted from the final layer of a pre-trained ResNet. The graph is constructed from a knowledge graph and each node is represented by a vector that encodes semantic class information, in our experiments the classes word embedding. The network consists of two phases, a descendant phase where each node receives knowledge form its descendants and an ancestor phase, where it receives knowledge from its ancestors.
  }\vspace{-1mm}
  \label{fig:arch}
\end{figure*}

%\subsection{Zero-shot learning}
Let $\mathcal{C}$ denote the set of all classes and $\mathcal{C}_{te}$ and $\mathcal{C}_{tr}$ the test and training classes, respectively. Further, assume that the training and test classes are disjoint $\mathcal{C}_{te} \cap \mathcal{C}_{tr} = \emptyset$ and that we are given a $S$ dimensional semantic representation vector $z\in \mathbb{R}^{S}$ for all classes and a set of training data points $\mathcal{D}_{tr}=\{(\vec{X}_i,c_i)\, i=1,..., N\}$, where $\vec{X}_i$ denotes the $i$-th training image and $c_i\in\mathcal{C}_{tr}$ the corresponding class label. In this setting, zero-shot classification aims to predict the class labels of a set of test data points to the set of classes $\mathcal{C}_{te}$. Note that, unlike traditional classification, the test data set points have to be assigned to previously unseen classes.
%Unlike the common approach of training classifiers $\mathcal{X} \rightarrow \mathcal{Y}$, where image examples for all classes $\mathcal{Y}$ are available, only image examples for some classes are available. With the ever-growing supply of image data, from an ever-expanding number of classes, it is of high importance to be able to perform zero-shot classification.

\subsection{Graph Convolutional Networks for Zero-Shot Learning}
In this work, we perform zero-shot classification by using the word embedding of the class labels and the knowledge graph to predict classifiers for each unknown class in form of last layer CNN weights.
Our zero-shot learning framework is illustrated in Fig.~\ref{fig:arch}. \comment{The last layer CNN weights are interpreted as a class-specific classifier for a given output class on top of the extracted CNN features. The zero-shot task can then be expressed as predicting a new set of weights for each of the unseen classes in order to extend the output layer of the CNN. Our {\ADGN} takes as input the combined knowledge graph for all seen and unseen classes, where each class is represented by a word embedding vector that encodes the class name. It is then trained to predict the last layer CNN weights for all (seen and unseen) classes in a semi-supervised manner. Exploiting the knowledge graph allows us to capture semantic relationships between classes, while the word embedding provides a semantic description of each specific class. During inference, the predicted weights can then be used to extend the set of output classes in the original CNN to enable classification of datapoints from unseen classes.}

More specifically, given a graph with $N$ nodes and $S$ input features per node, $X \in \mathbb{R}^{N \times S}$ denotes the feature matrix. Here each node represents one distinct concept/class in the classification task and each concept is represented by the word vector of the class name. 
The connections between the classes in the knowledge graph are encoded in form of a symmetric adjacency matrix $A \in \mathbb{R}^{N\times N}$, which also includes self-loops.
We employ a simple propagation rule to perform convolutions on the graph
\begin{equation}
    H^{(l+1)} = \sigma\left(D^{-1}A H^{(l)} \Theta^{(l)}\right) \;,
\label{eq:update_eq}
\end{equation}
where $H^{(l)}$ represents the activations in the $l^\text{th}$ layer and $\Theta \in \mathbb{R}^{S \times F}$ denotes the trainable weight matrix for layer $l$ with $F$ corresponding to the number of learned filters. For the first layer, $H^{(0)}=X$. $\sigma(\cdot)$ denotes a nonlinear activation function, in our case a Leaky ReLU. $D_{ii}=\sum_j A_{ij}$ is a degree matrix $D \in \mathbb{R}^{N\times N}$, which normalizes rows in $A$ to ensure that the scale of the feature representations is not modified by $A$. Similarly to previous work done on graph convolutional neural networks, this propagation rule can be interpreted as a spectral convolution~\cite{kipf2016semi}. 

%and also do not employ the symmetric normalization $D^{-1/2}AD^{-1/2}$ that was used in~\cite{wang2018zero} \commentY{Actually symmetric or not does not influence much, but currently the results are based on non-sym one. Should we mention that this modification is negligible? But Dense have to use non-sym.}. 
%See supplementary materials for a small comparison. \commentM{Can report some numbers in the supplementary material for this?}

The model is trained to predict the classifier weights for the seen classes by optimizing the loss
\begin{equation}
    \mathcal{L} = \frac{1}{2M} \sum_{i=1}^M \sum_{j=1}^P (W_{i,j} - \widetilde{W}_{i,j})^2 \;,
    \label{eq:loss}
\end{equation}
where ${\widetilde{W}}\in \mathbb{R}^{M \times P}$ denotes the prediction of the GCN for the known classes and therefore corresponds to the $M$ rows of the GCN output, which correspond to the training classes. 
$M$ denotes the number of training classes and $P$ denotes the dimensionality of the weight vectors. The ground truth weights are obtained by extracting the last layer weights of a pre-trained CNN and denoted as $W \in \mathbb{R}^{M \times P}$. 
During the inference phase, the features of new images are extracted from the CNN and the classifiers predicted by the GCN are used to classify the features.

\comment{However, the Laplacian smoothing operation in matrix form can be written as $(I-\gamma D^{-1}L)H$, as also noted in Li \etal~\cite{li2018deeper}. Substituting the graph Laplacian with its definition $L=D-A$ the operation simplifies for $\gamma=1$ (looking only at the immediate neighbors) to $D^{-1}AH$. This corresponds in parts to the graph convolution operation in Eq.~\ref{eq:update_eq}. 
%meaning that the complete update rule is thus a smoothing operating followed by a matrix multiplication with the weights $\Theta^{(l)}$, which can be interpreted as a fully connected layer.
Thus, repeatedly applying Eq.~\ref{eq:update_eq} in a multi-layer GCN architecture will lead to repeated Laplacian smoothing, thus diluting the information. Empirical evidence is provided in the model analysis section (Sec.~\ref{sec:model_analysis}).}

\subsection{Dense Graph Propagation Module}
Our {\ADGN} for zero-shot learning aims to use the hierarchical graph structure for the zero-shot learning task and avoids dilution of knowledge by intermediate nodes. This is achieved using a dense graph connectivity scheme consisting of two phases, namely descendant propagation and ancestor propagation. This two-phase approach further enables the model to learn separate relations between a node and its ancestors and a node and its descendants. Table~\ref{tab:two_phase} in the model analysis section provides empirical evidence for this choice. Unlike the GCN, we do not use the knowledge graph relations directly as an adjacency graph to include information from neighbors further away. We do therefore not suffer from the problem of knowledge being washed out due to averaging over the graph. Instead, we introduce two separate connectivity patterns, one where nodes are connected to all their ancestors and one where nodes are connected to all descendants. We use two adjacency matrices: $A_{a} \in \mathbb{R}^{N\times N}$ denotes the connections from nodes to their ancestors, whereas $A_{d}$ denotes the connections from nodes to their descendants. Note, as a given node is the descendant of its ancestors, the difference between the two adjacency matrices is a reversal of their edges $A_{d} = A_{a}^T$. Unlike previous approaches, this connectivity pattern allows nodes direct access to knowledge in their extended neighborhood as opposed to knowledge that has been modified by intermediate nodes. Note that both these adjacency matrices include self-loops. The connection pattern is illustrated in Fig.~\ref{fig:graph_connection}.
The same propagation rule as in Eq.~\ref{eq:update_eq} is applied consecutively for the two connectivity patterns leading to the overall {\ADGN} propagation rule
\begin{equation}
    H = \sigma\left(D_{a}^{-1} A_{a} \sigma\left(D_{d}^{-1} A_{d} X \Theta_{d} \right) \Theta_{a} \right) \;.
\label{eq:update_eq2}
\end{equation}

\textbf{Distance weighting scheme}
In order to allow {\ADGN} to weigh the contribution of various neighbors in the dense graph, we propose a weighting scheme that weighs a given node's neighbors based on the graph distance from the node. Note, the distance is computed on the knowledge graph and not the dense graph. We use $w^a=\{w_i^a\}_{i=0}^K$ and $w^d=\{w_i^d\}_{i=0}^K$ to denote the \commentM{learned} weights for the ancestor and the descendant propagation phase, respectively. $w_i^a$ and $w_i^d$ correspond to weights for nodes that are $i$ hops away from the given node. $w_0^a, w_0^d$ correspond to self-loops and $w_K^a, w_K^d$ correspond to the weights for \commentM{all} nodes further than $K-1$ hops away. We normalize the weights using a softmax function $\alpha_k^a = \text{softmax}(w_k^a)=\frac{\exp(w_k^a)}{\sum_{i=0}^K \exp(w_i^a)}$. Similarly, $\alpha_k^d=\text{softmax}(w_k^d)$. The weighted propagation rule in Eq.~\ref{eq:update_eq2} becomes
\begin{equation}
    H = \sigma\left(\sum_{k = 0}^K \alpha_k^a D_k^{{a}^{-1}} A_k^a \sigma\left(\sum_{k = 0}^K \alpha_k^d D_k^{{d}^{-1}} A_k^d X \Theta_{d} \right) \Theta_{a} \right)\;,
\label{eq:update_eq_attention}
\end{equation}
where $A_k^a$ and $A_k^d$ denote the parts of the adjacency matrices that only contain the $k$-hop edges for the ancestor and descendant propagation phase, respectively. $D_k^{{a}}$ and $D_k^{{d}}$ are the corresponding degree matrices for $A_k^a$ and $A_k^d$. \commentM{As weights are shared across the graph, the proposed weighting scheme only adds $2\times (K+1)$ parameters to the model, where $K$ tends to be small ($K=4$ in our experiments).}

Our proposed weighting scheme is related to the attention mechanisms in graph convolutional neural networks~\cite{velickovic2017graph}. However, unlike attention approaches, our weighting scheme adds only a negligible amount of parameters and does not add the potentially considerable memory overhead of attention approaches. % Delete potentially considerable?
Further, in our zero-shot learning setting, we observed a drop in performance when including \commentMG{the attention approach proposed in~\cite{velickovic2017graph}. We hypothesize that this is due to the fact that a more complex model will be more prone to overfit given the limited amount of labeled data (sparsely labeled graph). Results are provided in the supplementary material.} 
%attention approaches and we hypothesize that this is due to the fact that a more complex model will be more prone to overfit given the limited amount of labeled data (sparsely labeled graph). \commentMG{Results are provided in the supplementary material.}

\subsection{Finetuning}
Training is done in two stages, where the first stage trains the {\ADGN} to predict the last layer weights of a pre-trained CNN using Eq.~\ref{eq:loss}. Note, ${\widetilde{W}}$, in this case, contains the $M$ rows of $H$, which correspond to the training classes. In order to allow the feature representation of the CNN to adapt to the new class classifiers, we train the CNN by optimizing the cross-entropy classification loss on the seen classes in a second stage. During this stage, the last layer weights are fixed to the predicted weights of the training classes in the {\ADGN} and only the feature representation is updated. This can be viewed as using the {\ADGN} as a constraint for the CNN, as we indirectly incorporate the graph information to constrain the CNN output space.
%\commentM{Add more detail}

%%%%%%%%%%%%%%%%%%%%%%%%%%%%%%%%%%%%%%%%%%%%%%%%
%%%%%%%%%%%%%%%% Experiments %%%%%%%%%%%%%%%%%%%
%%%%%%%%%%%%%%%%%%%%%%%%%%%%%%%%%%%%%%%%%%%%%%%%

\section{Experiments}
We perform a comparative evaluation of the {\ADGN} against previous state-of-the-art on the ImageNet dataset~\cite{deng2009imagenet}, the largest commonly used dataset for zero-shot learning \footnote{\commentMG{Additional experiments have been performed on the AWA2 dataset and can be found in the supplementary material.}}. In our work, we follow the train/test split suggested by Frome \etal~\cite{frome2013devise}, who proposed to use the 21K ImageNet dataset for zero-shot evaluation. They define three tasks in increasing difficulty, denoted as "2-hops", "3-hops" and "All". Hops refer to the distance that classes are away from the ImageNet 2012 1K classes in the ImageNet hierarchy and thus is a measure of how far unseen classes are away from seen classes. "2-hops" contains all the classes within two hops from the seen classes and consists of roughly 1.5K classes, while "3-hops" contains about 7.8K classes. "All" contains close to 21K classes. None of the classes are contained in the ImageNet 2012 dataset, which was used to pre-train the ResNet-50 model.
Mirroring the experiment setup in~\cite{frome2013devise,norouzi2013zero,wang2018zero} we further evaluate the performance when training categories are included as potential labels. Note that since the only difference is the number of classes during the inference phase, the model does not have to be retrained. We denote the splits as "2-hops+1K", "3-hops+1K", "All+1K".

\subsection{Training details}
We use a ResNet-50~\cite{He2015} model that has been pre-trained on the ImageNet 2012 dataset. Following Wang \etal~\cite{wang2018zero}, we use the GloVe text model~\cite{pennington2014glove} trained on the Wikipedia dataset as the feature representation of our concepts in the graph. The {\ADGN} model consists of two layers as illustrated in Eq.~\ref{eq:update_eq2} with feature dimensions of $2048$ and the final output dimension corresponds to the number of weights in the last layer of the ResNet-50 architecture, $2049$ for weights and bias. Following the observation of Wang \etal~\cite{wang2018zero}, we perform L2-Normalization on the outputs as it regularizes the outputs into similar ranges. Similarly, we also normalize the ground truth weights produced by the CNN. We further make use of Dropout~\cite{srivastava2014dropout} with a dropout rate of $0.5$ in each layer. The model is trained for $3000$ epochs with a learning rate of $0.001$ and weight decay of $0.0005$ using Adam~\cite{kingma2014adam}. We make use of leaky ReLUs with a negative slope of $0.2$. The number of values per phase $K$ was set to $4$ as additional weights had diminishing returns. The proposed {\ADGN} model is implemented in PyTorch~\cite{paszke2017automatic} and training and testing are performed on a GTX 1080Ti GPU. Finetuning is done for 20 epochs using SGD with a learning rate of 0.0001 and momentum of 0.9.

%ZSL: 10% 2.29%, 0.94% for 3H and ALL respectively GZSL:  4.38% 1.26%, 0.48% for 3H and ALL 

\begin{table*}[tb]
\def\arraystretch{0.8}
\setlength{\tabcolsep}{4pt}
\small
    %\caption{Top-k accuracy for the different models on the ImageNet dataset.}
    \begin{minipage}[t]{.48\linewidth}
      \caption[t]{
      %Top-k accuracy for the different models on the ImageNet dataset. Accuracy when only testing on unseen classes. Results indicated with $^*$, $^{**}$, $^\dagger$, and $^\ddagger$ are taken from \cite{changpinyo2016synthesized}, \cite{xian2018feature}, \cite{changpinyo2017predicting}, and \cite{wang2018zero}, respectively.
      Top-k accuracy for the different models on the ImageNet dataset. Accuracy when only testing on unseen classes. Results indicated with $^*$, $^\dagger$, and $^\ddagger$ are taken from \cite{changpinyo2016synthesized}, \cite{changpinyo2017predicting}, and \cite{wang2018zero}, respectively.
      }
      \label{tab:results}
      \centering
        \begin{tabular}{l|c|ccccc}%c}
        %\cmidrule[1.5pt]{1-3}
        \toprule
        \bf \multirow{2}{*}{Test set} & \bf\multirow{2}{*}{Model} & \multicolumn{5}{c}{\bf Hit@k (\%)}\\
        & & 1 & 2 & 5 & 10 & 20\\
        \midrule
        {\multirow{7}{*}{\bf 2-hops}} & ConSE$^*$ & 8.3 & 12.9 & 21.8 & 30.9 & 41.7\\
        %& FGN$^{**}$ & 10.0 & - & - & - & - \\ 
        & SYNC$^*$ & 10.5 & 17.7 & 28.6 & 40.1 & 52.0\\
        & EXEM$^\dagger$ & 12.5 & 19.5 & 32.3 & 43.7 & 55.2\\
        & GCNZ$^\ddagger$ & 19.8 & 33.3 & 53.2 & 65.4 & 74.6\\
        %& baseline & 23.45 & 36.78 & 56.43 & 68.29 & 77.29\\
        \cline{2-7}
        %\Tstrut & 1-layer GCN(-f) & 24.8 & 38.3 & 57.5 & 69.9 & 79.6\\
        %& \ADGN(-wf) & 23.8 & 36.9 & 56.2 & 69.1 & 78.6\\
        %& \ADGN(-f) & 24.6 & 37.8 & 56.9 & 69.6 & 79.3\\
        \Tstrut & SGCN (ours) & 26.2 & 40.4 & 60.2 & 71.9 & 81.0\\
        %& \ADGN(-w) & 25.4 & 39.5 & 59.9 & 72.0 & 80.9\\
        & {\ADGN} (ours)& {\bf 26.6} & {\bf 40.7} & {\bf 60.3} & {\bf 72.3} & {\bf 81.3}\\
        \midrule
        {\multirow{7}{*}{\bf 3-hops}} & ConSE$^*$ & 2.6 & 4.1 & 7.3 & 11.1 & 16.4\\
        %& FGN$^{**}$ & 2.3 & - & - & - & - \\ 
        & SYNC$^*$ & 2.9 & 4.9 & 9.2 & 14.2 & 20.9\\
        & EXEM$^\dagger$ & 3.6 & 5.9 & 10.7 & 16.1 & 23.1\\
        & GCNZ$^\ddagger$ & 4.1 & 7.5 & 14.2 & 20.2 & 27.7\\
        %& baseline & 4.68 & 8.28 & 15.71 & 23.20 & 31.87\\
        \cline{2-7}
        \Tstrut & SGCN (ours) & 6.0 & 10.4 & 18.9 & 27.2 & 36.9\\
        %& \ADGN(-w) & 6.2 & 10.5 & 19.2 & {\bf 27.7} & {\bf 37.7}\\
        & {\ADGN} (ours) & {\bf 6.3} & {\bf 10.7} & {\bf 19.3} & {\bf 27.7} & {\bf 37.7} \\
        \midrule
        {\multirow{7}{*}{\bf All}} & ConSE$^*$ & 1.3 & 2.1 & 3.8 & 5.8 & 8.7\\
        %& FGN$^{**}$ & 0.9 & - & - & - & - \\ 
        & SYNC$^*$ & 1.4 & 2.4 & 4.5 & 7.1 & 10.9\\
        & EXEM$^\dagger$ & 1.8 & 2.9 & 5.3 & 8.2 & 12.2\\
        & GCNZ$^\ddagger$ & 1.8 & 3.3 & 6.3 &  9.1 & 12.7\\
        \cline{2-7}
        %& baseline & 2.04 & 3.59 & 6.96 & 10.66 & 15.39\\
        \Tstrut & SGCN (ours) & 2.8 & 4.9 & 9.1 & 13.5 & 19.3\\
        %& \ADGN(-w) & 2.9 & {\bf 5.0} & {\bf 9.3} & {\bf 13.9} & {\bf 19.9}\\
        & {\ADGN} (ours) & {\bf 3.0} & {\bf 5.0} & {\bf 9.3} & {\bf 13.9} & {\bf 19.8} \\
        \bottomrule
        \end{tabular}
    \end{minipage}%
    \hspace{0.5em}
    %\small
    \begin{minipage}[t]{.49\linewidth}
    
      \centering
        \caption[t]{Top-k accuracy for the different models on the ImageNet dataset. Accuracy when testing on seen and unseen classes. Results indicated with $^{\dagger\dagger}$, $^{\ddagger\ddagger}$, and $^\ddagger$ are taken from \cite{frome2013devise}, \cite{norouzi2013zero}, and \cite{wang2018zero}, respectively.}
        \label{tab:results_generalized}
        \begin{tabular}{l|c|ccccc}%c}
        %\cmidrule[1.5pt]{1-3}
        \toprule
        \bf \multirow{2}{*}{Test set} & \bf\multirow{2}{*}{Model} & \multicolumn{5}{c}{\bf Hit@k (\%)}\\
        & & 1 & 2 & 5 & 10 & 20\\
        \midrule
        {\multirow{7}{*}{\bf 2-hops+1K}} & DeViSE$^{\dagger\dagger}$ & 0.8 & 2.7 & 7.9 & 14.2 & 22.7\\
        & ConSE$^{\ddagger\ddagger}$ & 0.3 & 6.2 & 17.0 & 24.9 & 33.5\\
        & ConSE$^\ddagger$ & 0.1 & 11.2 & 24.3 & 29.1 & 32.7\\
        %& FGN$^{**}$ & 4.4 & - & - & - & - \\ 
        & GCNZ$^\ddagger$ & 9.7 & 20.4 & 42.6 & 57.0 & 68.2\\
        \cline{2-7}
        %& baseline & 23.45 & 36.78 & 56.43 & 68.29 & 77.29\\
        %\Tstrut & 1-layer(-f) GCN & 11.4 & 25.2 & 47.7 & 61.8 & 73.3\\
        %& \ADGN(-wf) & 9.6 & 22.5 & 44.9 & 60.6 & 72.3\\
        %& \ADGN(-f) & 10.6 & 24.1 & 46.2 & 61.4 & 73.1\\
        \Tstrut & SGCN (ours) & {\bf 11.9} & {\bf 27.0} & {\bf 50.8} & 65.1 & 75.9 \\
        %& \ADGN(-w) & 10.6 & 25.1 & 49.2 & 64.6 & 75.5\\
        & {\ADGN} (ours) & 10.3 & 26.4 & 50.3 & {\bf 65.2} & {\bf 76.0}\\
        \midrule
        {\multirow{7}{*}{\bf 3-hops+1K}} & DeViSE$^{\dagger\dagger}$ & 0.5 & 1.4 & 3.4 & 5.9 & 9.7\\
        & ConSE$^{\ddagger\ddagger}$ & 0.2 & 2.2 & 5.9 & 9.7 & 14.3\\
        & ConSE$^\ddagger$ & 0.2 & 3.2 & 7.3 & 10.0 & 12.2\\
        %& FGN$^{**}$ & 1.3 & - & - & - & - \\
        & GCNZ$^\ddagger$ & 2.2 & 5.1 & 11.9 & 18.0 & 25.6\\
        \cline{2-7}
        %& baseline & 23.45 & 36.78 & 56.43 & 68.29 & 77.29\\
        \Tstrut & SGCN (ours) & {\bf 3.2} & {\bf 7.1} & {\bf 16.1} & 24.6 & 34.6\\
        %& \ADGN(-w) & 3.0 & 6.9 & 15.8 & 24.6 & 35.0\\
        & {\ADGN} (ours) & 2.9 & {\bf 7.1} & {\bf 16.1} & {\bf 24.9} & {\bf 35.1}\\
        \midrule
        {\multirow{7}{*}{\bf All+1K}} & DeViSE$^{\dagger\dagger}$ & 0.3 & 0.8 & 1.9 & 3.2 & 5.3\\
        & ConSE$^{\ddagger\ddagger}$ & 0.2 & 1.2 & 3.0 & 5.0 & 7.5\\
        & ConSE$^\ddagger$ & 0.1 & 1.5 & 3.5 & 4.9 & 6.2\\
        %& FGN$^{**}$ & 0.5 & - & - & - & - \\ 
        & GCNZ$^\ddagger$ & 1.0 & 2.3 & 5.3 & 8.1 & 11.7\\
        %& baseline & 23.45 & 36.78 & 56.43 & 68.29 & 77.29\\
        \cline{2-7}
        \Tstrut & SGCN (ours) & {\bf 1.5} & {\bf 3.4} & 7.8 & 12.3 & 18.2\\
        %& \ADGN(-w) & {\bf 1.5} & {\bf 3.4} & 7.8 & 12.4 & 18.6\\
        & {\ADGN} (ours) & 1.4 & {\bf 3.4} & {\bf 7.9} & {\bf 12.6} & {\bf 18.7}\\
        \bottomrule
        \end{tabular}
    \end{minipage} 
\end{table*}

\subsection{Comparing approaches}
We compare our {\ADGN} to the following approaches:
{\bf Devise}~\cite{frome2013devise} linearly maps visual information in form of features extracted by a convolutional neural network to the semantic word-embedding space. The transformation is learned using a hinge ranking loss. Classification is performed by assigning the visual features to the class of the nearest word-embedding.
{\bf ConSE}~\cite{norouzi2013zero} projects image features into a semantic word embedding space as a convex combination of the $T$ closest seen classes semantic embedding weighted by the probabilities that the image belongs to the seen classes. The probabilities are predicted using a pre-trained convolutional classifier. Similar to Devise, ConSE assigns images to the nearest classes in the embedding space.
{\bf EXEM}~\cite{changpinyo2017predicting} creates visual class exemplars by averaging the PCA projections of images belonging to the same seen class. A kernel-based regressor is then learned to map a semantic embedding vector to the class exemplar. For zero-shot learning visual exemplars can be predicted for the unseen classes using the learned regressor and images can be assigned using nearest neighbor classification.
{\bf SYNC}~\cite{changpinyo2016synthesized} aligns a semantic space (e.g., the word-embedding space) with a visual model space, adds a set of phantom object classes in order to connect seen and unseen classes, and derives new embeddings as a convex combination of these phantom classes.
%\commentMG{{\bf FGN}~\cite{xian2018feature} uses a generative adversarial network to produce CNN features for the unseen classes conditioned on the semantic information.}
{\bf GCNZ}~\cite{wang2018zero} represents the current state of the art and is the approach most related to our proposed {\ADGN}. A GCN is trained to predict last layer weights of a convolutional neural network.

Guided by experimental evidence (see our analysis in Table~\ref{tab:layer_size} in the model analysis section) and our intuition that extensive smoothing is a disadvantage for the weight regression in zero-shot learning, we add a single-hidden-layer GCN ({\bf SGCN}) with non-symmetric normalization ($D^{-1}A$) (as defined in Eq.~\ref{eq:update_eq}) as another baseline. Note, GCNZ made use of a symmetric normalization ($D^{-1/2}AD^{-1/2}$) but our experimental evaluation indicates that the difference is negligible. For the interested reader, an analysis of the effect of the changes between GCN and SGCN is included in the supplementary material. SGCN further yields a better baseline since our proposed {\ADGN} also utilizes the non-symmetric normalization. As {\ADGN}, our SGCN model makes use of the proposed two-stage finetuning approach. 

\subsection{Comparison to state-of-the-art methods}

\begin{figure*}[tbp]
  \centering
  \includegraphics[trim={16.5cm 7.2cm 18.7cm 4.8cm},clip, width=0.95\linewidth]{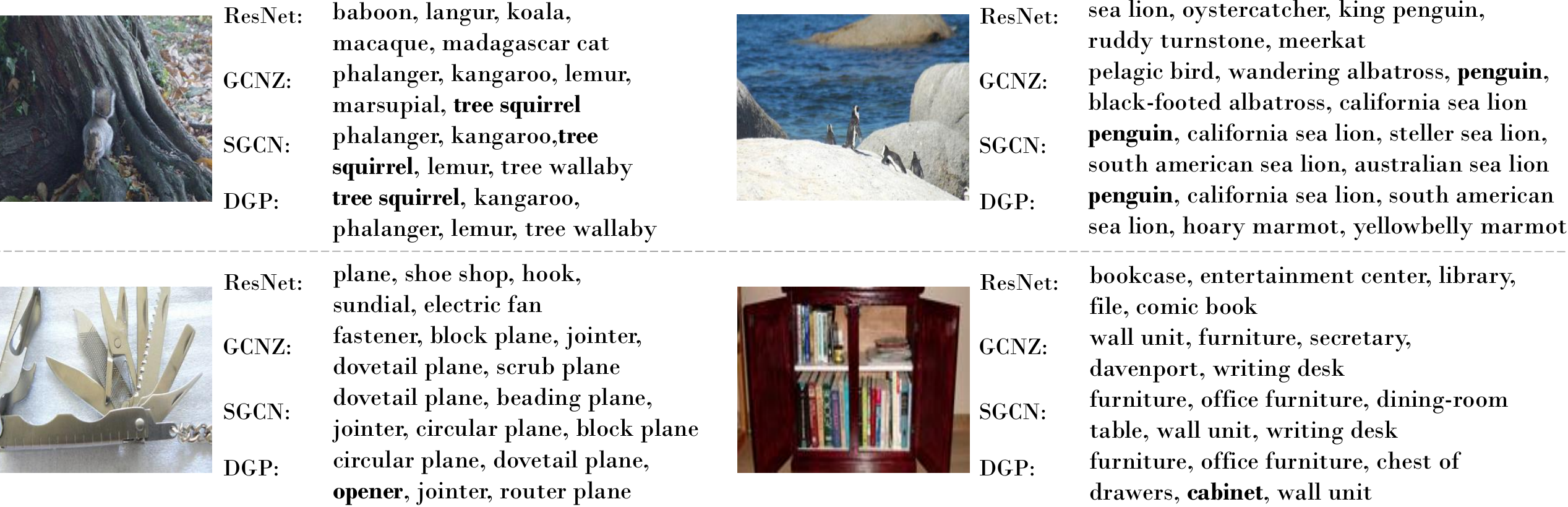}
  \caption{Qualitative result comparison. The correct class is highlighted in bold. We report the top-5 classification results.}
  \label{fig:qualiRes}
\end{figure*}

\textbf{Quantitative results} for the comparison on the ImageNet datasets are shown in Table~\ref{tab:results}. Compared to previous results such as ConSE~\cite{changpinyo2016synthesized}, EXEM~\cite{changpinyo2017predicting}, and GCNZ~\cite{wang2018zero} our proposed methods outperform the previous results with a considerable margin, achieving, for instance, more than 50\% relative improvement for Top-1 accuracy on the 21K ImageNet "All" dataset. We observe that our methods especially outperform the baseline models on the "All" task, illustrating the potential of our methods to more efficiently propagate knowledge. {\ADGN} also achieves consistent improvements over the SGCN model. We observed that finetuning consistently improved performance for both models in all our experiments. Ablation studies that highlight the impact of finetuning and weighting of neighbors for the 2-hop scenario can be found in Table~\ref{tab:resultsAblation}.  {\ADGN(-wf)} is used to denote the accuracy that is achieved after training the {\ADGN} model without weighting (adding no weights in Eq.~\ref{eq:update_eq_attention}) and without finetuning. \ADGN(-w) and \ADGN(-f) are used to denote the results for {\ADGN} without weighting and {\ADGN} without finetuning, respectively. We further report the accuracy achieved by the SGCN model without finetuning ({SGCN(-f)}). We observe that the proposed weighting scheme, which allows distant neighbors to have less impact, is crucial for the dense approach. Further, finetuning the model consistently leads to improved results.

%and perform ablation studies. {\ADGN(-wf)} is used to denote the accuracy that is achieved after training the {\ADGN} model without weighting and finetuning. \ADGN(-w) and \ADGN(-f) are used to denote the results for {\ADGN} without weighting and {\ADGN} without finetuning, respectively. We further report the accuracy achieved by the 1-layer GCN model and a finetuned version of it {1-layer GCN(+f)}. Compared to previous results such as ConSE~\cite{changpinyo2016synthesized}, EXEM~\cite{changpinyo2017predicting}, and GCNZ~\cite{wang2018zero} our proposed methods outperform the previous results with a considerable margin, achieving, for instance, more than 50\% relative improvement for Top-1 accuracy on the 21K ImageNet All dataset. 
%We observe that our methods especially outperform the baseline models on the "All" task, illustrating the potential of our methods to more efficiently propagate knowledge. Furthermore, we perform an ablation experiment for the 2-hops dataset where we analyze the effect of removing the weighting and finetuning of the model and observe that this consistently leads to a drop in model performance.
%Results further illustrate that the proposed {\ADGN}, which learns both feature extraction and the zero-shot classifiers, achieves better performance than all variants, demonstrating that our dense weighting scheme can make features more general and incorporates more graph information.
%, consistently leads to improvements and that {\ADGN} generally is able to outperform {\GN} as the model is more expressive.

\textbf{Qualitative results} of {\ADGN} and the SGCN are shown in Fig.~\ref{fig:qualiRes}. Example images from unseen test classes are displayed and we compare the results of our proposed {\ADGN} and the SGCN to results produced by a pre-trained ResNet. Note, ResNet can only predict training classes while the others predict classes not seen in training. For comparison, we also provide results for our re-implementation of GCNZ. We observe that the SGCN and {\ADGN} generally provide coherent top-5 results. All methods struggle to predict the \emph{opener} and tend to predict some type of \emph{plane} instead, however, {\ADGN} does include \emph{opener} in the top-5 results. We further observe that the prediction task on this dataset for zero-shot learning is difficult as it contains classes of fine granularity, such as many different types of squirrels, planes, and furniture. Additional examples are provided in the supplementary material.

\textbf{Testing including training classifiers.} Following the example of \cite{frome2013devise, norouzi2013zero, wang2018zero}, we also report the results when including both training labels and testing labels as potential labels during classification of the zero-shot examples. Results are shown in Table~\ref{tab:results_generalized}. For the baselines, we include two implementations of ConSE, one that uses AlexNet as a backbone~\cite{norouzi2013zero} and one that uses ResNet-50~\cite{wang2018zero}. Compared to Table~\ref{tab:results}, we observe that the accuracy is considerably lower, but the SGCN and {\ADGN} still outperform the previous state-of-the-art approach GCNZ. SGCN outperforms {\ADGN} for low $k$ in the Top-$k$ accuracy measure especially for the 2-hops setting, while {\ADGN} outperforms SGCN for larger $k$. We observe that {\ADGN} tends to favor prediction to the closest training classes for its Top-1 prediction (see Table~\ref{tab:results_seen_image}). However, this is not necessarily a drawback and is a well-known tradeoff~\cite{chao2016empirical} between performing well on the unseen classes and the seen classes, which are not considered in this setting. 
%In the next paragraph, we will evaluate the model's performance on the seen classes. 
This tradeoff can be controlled by including a novelty detector, which predicts if an image comes from the seen or unseen classes as done in \cite{socher2013zero} and then assign it to the zero-shot classifier or a classifier trained on the seen classes. Another approach is calibrated stacking~\cite{chao2016empirical}, which rescales the prediction scores of the known classes.

\begin{table}[tbp]
\def\arraystretch{0.8}
\setlength{\tabcolsep}{4pt}
\small
      \caption[t]{Results of the ablation experiments on the 2-hops dataset. (-f), (-w), and (-wf) indicate models without finetuning, weighting and without both weighting and finetuning, respectively.}
      \label{tab:resultsAblation}
      \centering
        \begin{tabular}{l|c|ccccc}%c}
        %\cmidrule[1.5pt]{1-3}
        \toprule
        \bf \multirow{2}{*}{Test set} & \bf\multirow{2}{*}{Model} & \multicolumn{5}{c}{\bf Hit@k (\%)}\\
        & & 1 & 2 & 5 & 10 & 20\\
        \midrule
        {\multirow{6}{*}{\bf 2-hops}} & SGCN(-f) & 24.8 & 38.3 & 57.5 & 69.9 & 79.6\\
        & \ADGN(-wf) & 23.8 & 36.9 & 56.2 & 69.1 & 78.6\\
        & \ADGN(-f) & 24.6 & 37.8 & 56.9 & 69.6 & 79.3\\
        & \ADGN(-w) & 25.4 & 39.5 & 59.9 & 72.0 & 80.9\\
        \cline{2-7}
        \Tstrut & SGCN (ours) & 26.2 & 40.4 & 60.2 & 71.9 & 81.0\\
        & {\ADGN} (ours) & {\bf 26.6} & {\bf 40.7} & {\bf 60.3} & {\bf 72.3} & {\bf 81.3}\\
        \bottomrule
        \end{tabular}
\end{table}

%\textbf{Study of domain shift issue.} 
%Zero-shot learning models should perform well not only on unseen but also on seen classes. 
To put the zero-shot performance into perspective, we perform experiments where we analyze how the model's performance on the original $1000$ seen classes is affected by domain shift as additional unseen classes (all 2-hop classes) are introduced. Table~\ref{tab:results_seen_image} shows the results when the model is tested on the validation dataset from ImageNet 2012. We compare the performance to our re-implementation of the GCNZ model with ResNet-50 backbone and also the performance from the original ResNet-50 model, which is trained only on the seen classes. It can be observed that both our methods outperform GCNZ.

\subsection{Model analysis}
\label{sec:model_analysis}
\textbf{Analysis of weighting scheme.}
To validate our intuition that weighting allows our approach to weigh distant neighbors less, we inspect the learned weights. For the first stage the weights are $0.244$, $0.476$, $0.162$, $0.060$, $0.058$ and for the second (final) stage they are $0.493$, $0.322$, $0.097$, $0.047$, $0.041$. Note, the first value corresponds to self-weighting, the second to the 1-hop neighbors, and so forth. It can be observed, that ancestors aggregate information mainly from their immediate descendants in the first phase and later distribute it to their descendants in the second phase. Further, we observe that distant neighbors have far less impact in the final stage. \commentM{This means that the model learns to preserve the overall graph structure imposed by the knowledge graph, where importance is governed by the distance in the graph.}
%\commentM{Table~\ref{tab:Attention} compares our weighting scheme to a more commonly used attention scheme, where a specific weight is learned for each individual node by ...\commentM{Add details}. We note, that our depth based weighting scheme outperforms the node-wise attention.\commentM{Other split results in appendix if possible}}. \commentM{Hypothesis: If we allow too much flexibility we loose the advantage of the graph model as it can learn to fit the training dataset well. Need to find a sweet spot ... depth based weighting?}

\begin{table}[tbp]
\setlength{\tabcolsep}{10pt}
\def\arraystretch{0.8}
\small
     \centering
        \caption{Performance on the seen ImageNet classes. ResNet represents ideal performance as it only predicts known classes. \commentMG{GCNZ is our reimplementation of~\cite{wang2018zero}.}}
        \label{tab:results_seen_image}
        \begin{tabular}{l|cccc}%c}
        %\cmidrule[1.5pt]{1-3}
        \toprule
        \bf \multirow{2}{*}{Model} &  \multicolumn{4}{c}{\bf Hit@k (\%)}\\
        & 1 & 2 & 5 & 10\\
        \midrule
        ResNet & 75.1 & 85.5 & 92.7 & 95.7\\
        %DeViSE~\cite{norouzi2013zero} & 53.2 & 65.2 & 76.7 & 83.3\\
        %ConSE~\cite{norouzi2013zero} & 54.3 & 61.9 & 68.0 & 71.6\\
        GCNZ & 38.3 & 62.9 & 82.3 & 89.8\\
        \midrule
        %1-layer GCN & 44.8 & 63.8 & 80.1 & 87.0\\
        %\ADGN(-wf) & 48.7 & 63.6 & 78.8 & 85.6\\
        %\ADGN(-f) & 47.0 & 63.5 & 79.1 & 86.0\\
        SGCN (ours) & 49.1 & 68.7 & {\bf 83.9} & {\bf 89.4}\\
        %\ADGN(-w) & 53.0 & 68.5 & 83.7 & 89.2\\
        {\ADGN} (ours) & {\bf 54.6} & {\bf 69.7} & 83.8 & 89.1\\
        \bottomrule
        \end{tabular}
\end{table}

\textbf{Analysis of number of layers.} 
We perform an empirical evaluation to verify that our intuition is correct and that additional hidden layers indeed cause a drop in performance when employing a GCN. Table~\ref{tab:layer_size} illustrates the performance when adding additional layers to the GCN for the 2-hops experiment. These results are reported without finetuning the model. In order to perform this ablation study we fix all hidden layers to have a dimensionality of 2048 with 0.5 dropout. \comment{We want to stress that there is a fundamental difference in our experimental setting and the study in Wang \etal~\cite{wang2018zero}, as their ablation study does not only consider a different number of layers in the network but also a different number of neurons per layer at the same time.} %where the number of hidden units is varied in addition to the number of layers.

\begin{table}[tbp]
\def\arraystretch{0.8}
\setlength{\tabcolsep}{8pt}
\small
 \caption{Results for 2-hops for SGCN without finetuning when increasing the depth.}
      \label{tab:layer_size}
      \centering
        \begin{tabular}{l|ccccc}%c}
        %\cmidrule[1.5pt]{1-3}
        \toprule
        \bf \#Layers & \multicolumn{5}{c}{\bf Hit@k (\%)}\\
        & 1 & 2 & 5 & 10 & 20\\
        \midrule
        1 & {\bf 24.8} & {\bf 38.3} & {\bf 57.5} & {\bf 69.9} & {\bf 79.6}\\
        2 & 24.2 & 37.7 & 57.4 & 69.2 & 78.1\\
        3 & 23.9 & 37.5 & 57.1 & 68.4 & 77.2\\
        %4 & 23.8 & 37.3 & 56.5 & 68.2 & 77.2\\
        \bottomrule
    \end{tabular}
    \vspace{0.5em}
    \end{table}

\begin{table}[tbp]
\def\arraystretch{0.8}
\setlength{\tabcolsep}{8pt}
\small
%\footnotesize
    %\caption{Top-k accuracy for the different models on the ImageNet dataset.}
    \caption{Results for 2-hops with/without separating the adjacency matrix into ancestors and descendants for DGP.}
        \label{tab:two_phase}
        \centering
        \begin{tabular}{l|ccccc}%c}
        %\cmidrule[1.5pt]{1-3}
        \toprule
        \bf Model & \multicolumn{5}{c}{\bf Hit@k (\%)}\\
        & 1 & 2 & 5 & 10 & 20\\
        \midrule
        without & 26.0 & 40.2 & 59.8 & 71.4 & 80.3\\
        with & {\bf 26.6} & {\bf 40.7} & {\bf 60.3} & {\bf 72.3} & {\bf 81.3}\\
        \bottomrule
        \end{tabular}
\end{table}
    
\textbf{Analysis of two-phase propagation.}
We further, perform an ablation study to analyze the benefit of a two-phase directed propagation rule where ancestors and descendants are considered individually. We compared this to two consecutive updates using the full adjacency matrix in the dense method and illustrate the results in Table~\ref{tab:two_phase}. Consistent improvements are obtained using our proposed two-phase directed propagation rule. 

\textbf{Robustness of results.}
Table~\ref{tab:significant} shows the mean and standard deviation for 3 runs for the 2-hops and All datasets. The results are stable over multiple runs and it can clearly be observed that as the number of classes increases (2-hops to all), results become more stable.

\begin{table}[tbp]
\def\arraystretch{0.8}
\setlength{\tabcolsep}{8pt}
\small
        \caption{Mean and standard deviation for 3 runs. More stable as the number of class increases.}
        \label{tab:significant}
        \centering
        \begin{tabular}{l|c|ccccc}%c}
        %\cmidrule[1.5pt]{1-3}
        \toprule
        \bf Test set & \bf Model & \multicolumn{2}{c}{\bf Hit@k (\%)}\\
        & & 1 & 2\\
        \midrule
        {\multirow{2}{*}{\bf 2-hops}} & SGCN & 26.17$\pm$0.03 & 40.41$\pm$0.03\\
        & DGP & 26.67$\pm$0.09 & 40.74$\pm$0.04\\
        \midrule
        {\multirow{2}{*}{\bf All}} & SGCN & 2.80$\pm$0.01 & 4.90$\pm$0.01\\
        & DGP & 2.95$\pm$0.00 & 5.05$\pm$0.02\\
        \bottomrule
        \end{tabular}
\end{table}

\textbf{Scalability.} 
To obtain good scalability it is important that the adjacency matrix $A$ is a sparse matrix so that the complexity of computing $D^{-1}AX\Theta$ is linearly proportional to the number of edges present in $A$. Our approach exploits the structure of knowledge graphs, where entities only have few ancestors and descendants, to ensure this. The adjacency matrix for the ImageNet hierarchy used in our experiments, for instance, has a density of $9.3\times 10^{-5}$, while our dense connections only increase the density of the adjacency matrix to $19.1\times 10^{-5}$. 

With regards to the number of parameters, the SGCN consists of 4,810,752 weights. {\ADGN} increases the number of trainable parameters by adding $2\times (K+1)$ additional weights. However, as $K=4$ in our experiments, this difference in the number of parameters is negligible. Overall the number of trainable parameters is considerably lower than that in the GCNZ model (9,527,808 weights).

%%%%%%%%%%%%%%%%%%%%%%%%%%%%%%%%%%%%%%%%%%%%%%%%
%%%%%%%%%%%%%%%%% Conclusion %%%%%%%%%%%%%%%%%%%
%%%%%%%%%%%%%%%%%%%%%%%%%%%%%%%%%%%%%%%%%%%%%%%%
\section{Conclusion}
In contrast to previous approaches using graph convolutional neural networks for zero-shot learning, we illustrate that the task of zero-shot learning benefits from shallow networks. Further, to avoid the lack of information propagation between distant nodes in shallow models, we propose {\ADGN}, which exploits the hierarchical structure of the knowledge graph by adding a weighted dense connection scheme.
Experiments illustrate the ability of the proposed methods, outperforming previous state-of-the-art methods for zero-shot learning.
In future work, we aim to investigate the potential of more advanced weighting mechanisms to further improve the performance of {\ADGN} compared to the SGCN. The inclusion of additional semantic information for settings where these are available for a subset of nodes is another future direction.

\small
\paragraph{Acknowledgments:}
This work was partially funded by the Norwegian Research Council FRIPRO grant no.\ 239844.

\cleardoublepage
{\small
\bibliographystyle{ieee}
\bibliography{egbib}
}

\cleardoublepage
%%%%%%%%% TITLE
%\title{%
%  \Large Rethinking Knowledge Graph Propagation for Zero-Shot Learning \\
%  
\section{Supplementary Materials}

%s\thispagestyle{empty}

%%%%%%%%% ABSTRACT

\subsection{Additional Qualitative Examples}
\begin{figure*}[t]
%\centering
\includegraphics[width=1\linewidth]{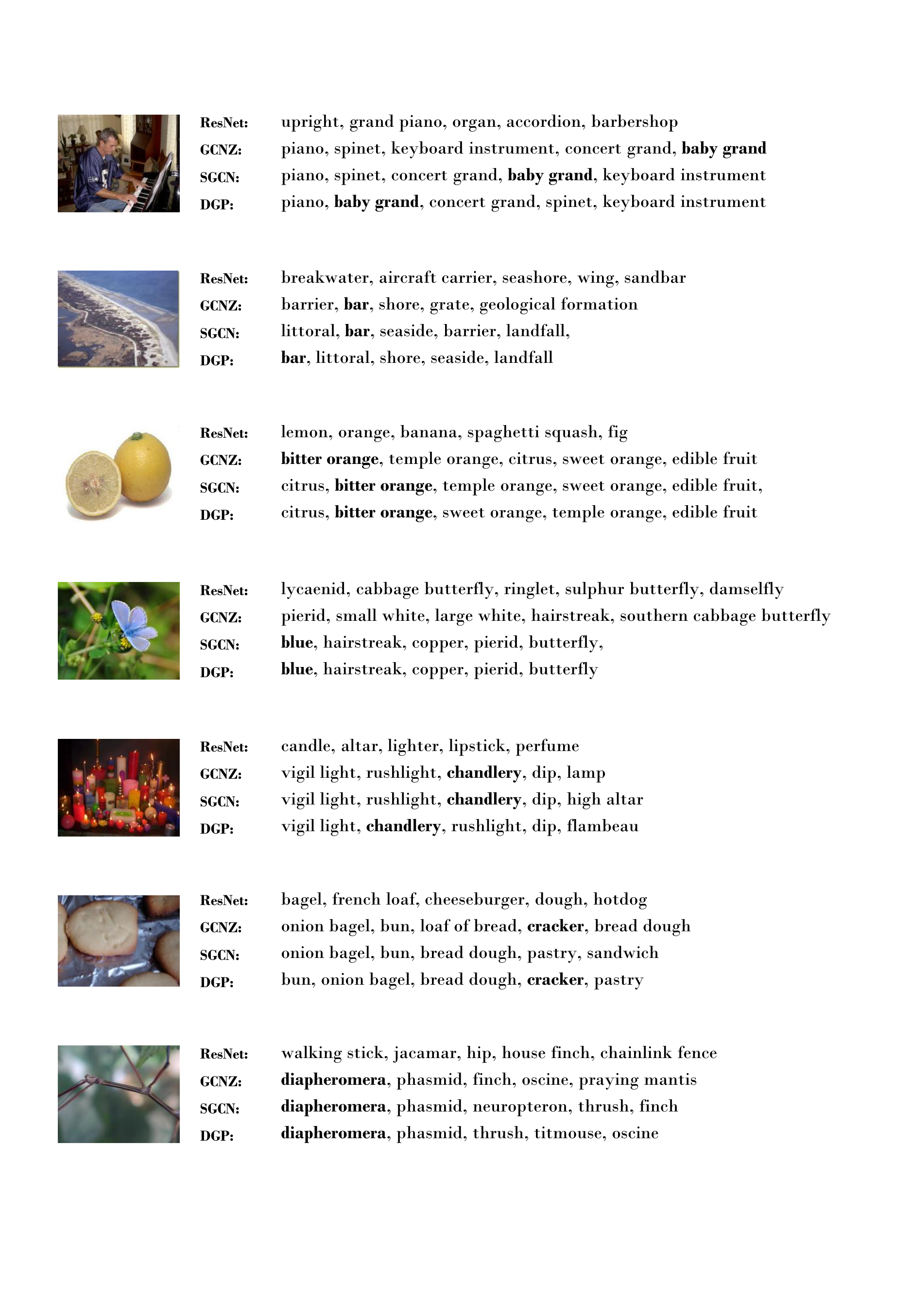}
\caption{Qualitative result comparison. The correct class is highlighted in bold. We report the top-5 classification results.}
\label{fig:qualiRes1}
\end{figure*}

\begin{figure*}[t]
%\centering
\includegraphics[width=1\linewidth]{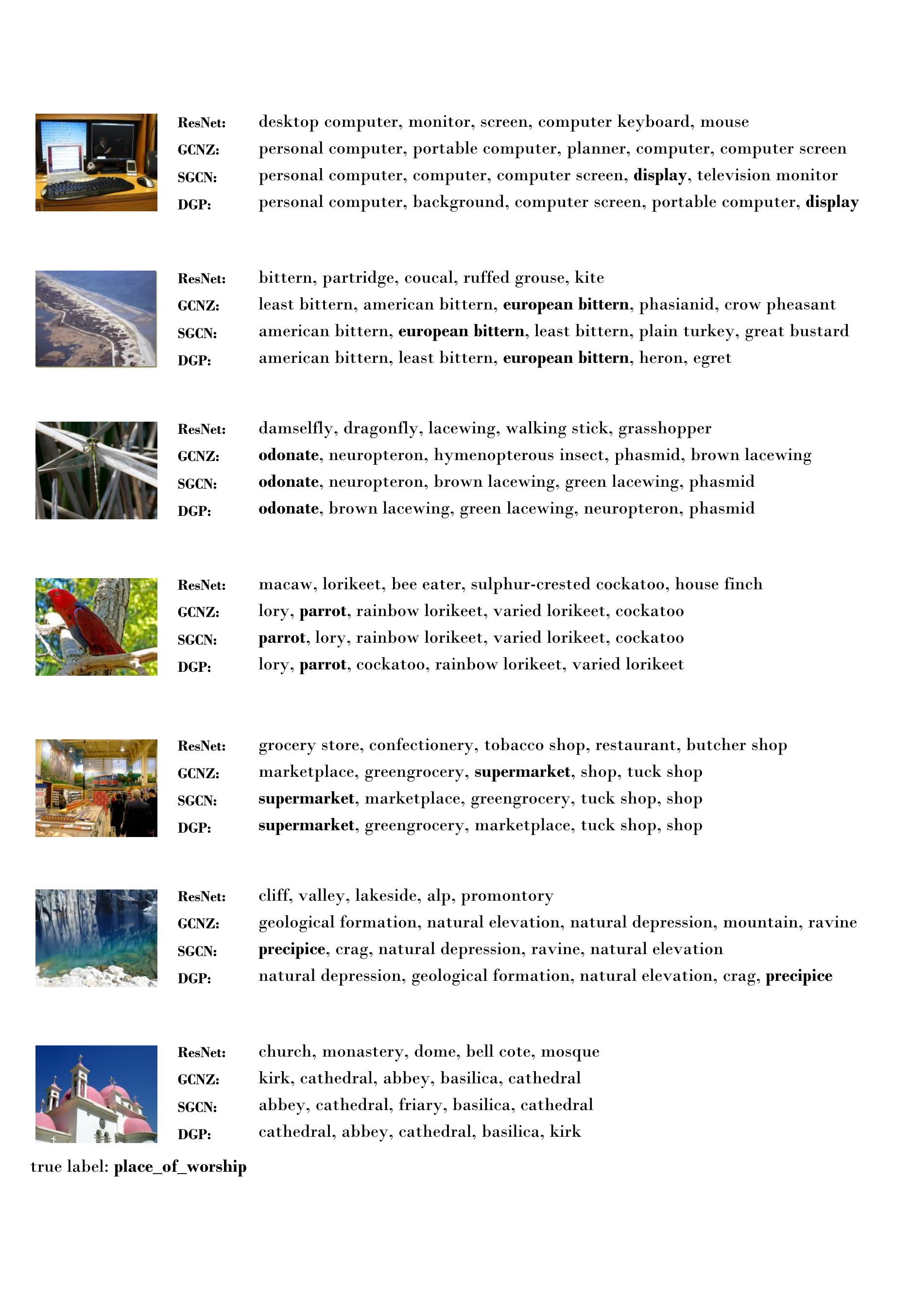}
\caption{Qualitative result comparison. The correct class is highlighted in bold. We report the top-5 classification results.}
\label{fig:qualiRes2}
\end{figure*}

Figure~\ref{fig:qualiRes1} and ~\ref{fig:qualiRes2} provide further qualitative results of our single-hidden-layer GCN (SGCN) and Dense Graph Propagation Module~({\ADGN}) compared to a standard ResNet and GCNZ, our reimplementation of~\cite{wang2018zero}.

\subsection{Performance improvements between GCNZ and SGCN}
\label{sec:xiaolongComparison}

Table~\ref{tab:xiaolongComparison} explains the performance difference between our SGCN, our reimplementation of GCNZ and the reported results in \cite{wang2018zero}. Note, unless otherwise stated training is performed for $3000$ epochs. Non-symmetric normalization ($D^{-1}A$) is denoted as \emph{non-sym} in the normalization column, while a symmetric normalization ($D^{-1/2}AD^{-1/2}$) is denoted as \emph{sym}. No finetuning has been performed for SGCN in these results.

\begin{table}[h]
    %\caption{Top-k accuracy for the different models on the ImageNet dataset.}
    \setlength{\tabcolsep}{2pt}
    \caption{Illustration of the improvements between the original results of GCNZ in \cite{wang2018zero}, our reimplementation of GCNZ and our SGCN. GCNZ$^\dagger$ corresponds to updated results from \cite{wang2018zero} (taken from \url{https://github.com/JudyYe/zero-shot-gcn}). GCNZ$^\ddagger$ is our re-implementation of \cite{wang2018zero}.}
        \label{tab:xiaolongComparison}
        \centering
        \begin{tabular}{l|cccccc}%c}
        %\cmidrule[1.5pt]{1-3}
        \toprule
        \bf Model & Norm & \multicolumn{5}{c}{\bf Hit@k (\%)}\\
        & & 1 & 2 & 5 & 10 & 20\\
        \midrule
        GCNZ (300 epochs) \cite{wang2018zero} & sym & 19.8 & 33.3 & 53.2 & 65.4 & 74.6\\
        GCNZ$^\dagger$ (300 epochs) \cite{wang2018zero} & sym & 21.0 & 33.7 & 52.7 & 64.8 & 74.3\\
        GCNZ$^\ddagger$ (300 epochs) & sym & 21.4 & 34.7 & 54.3 & 67.5 & 77.6\\
        GCNZ$^\ddagger$ & sym & 23.5 & 36.9 & 56.5 & 68.8 & 78.0\\
        SGCN (ours) & sym & 24.6 & 38.1 & 57.6 & 70.0 & 79.7\\
        SGCN (ours) & non-sym & 24.8 & 38.3 & 57.5 & 69.9 & 79.6\\
        \bottomrule
        \end{tabular}
\end{table}

\subsection{Performance on AWA2}

AWA2 is a replacement for the original AWA dataset and represents more traditional zero-shot learning datasets, where most approaches rely on class-attribute information. It consists of 50 animal classes, with a total of 37,322 images and an average of 746 per class.
The dataset further consists of 85-attribute features per class. We report results on the proposed split in~\cite{xian2018zero} to ensure that there is no overlap between the test classes and the ImageNet 2012 dataset. In the proposed split, 40 classes are used for training and 10 for testing. 
AWA2 test classes are contained in the 21K ImageNet classes and several of the training classes (24 out of 40) that are in the proposed split overlap with the ImageNet 2012 dataset. We, therefore, use a unified approach for both datasets.

Results for the AWA2 dataset are presented in Table~\ref{tab:results_awa}. Note that our model differs considerably from the baselines as it does not make use of the attributes provided in the dataset. To illustrate the merits of our approach, we re-implement~\cite{wang2018zero}, as it represents the method which is closest related to our approach and also makes use of word embeddings and a knowledge graph. We observe that our methods also outperforms~\cite{wang2018zero}, however, the improvement is lower than on the ImageNet dataset, which we believe is due to the arguably simpler task with the number of classes being considerably lower. Note, all methods, except SYNC, use a pretrained network trained on the 1K ImageNet classes. GCNZ and our DGP do not make use of the attribute information supplied for AWA2, however, both methods use the ImageNet knowledge graph. 

\begin{table}[h]
\centering
\def\arraystretch{0.6}
\caption{Top-1 accuracy results for unseen classes on AWA2. Results for ConSE, Devise and SYNC obtained from \cite{xian2018zero}.}
\label{tab:results_awa}
\begin{tabular}{l|c}%c}
%\cmidrule[1.5pt]{1-3}
\toprule
\bf Model &  \bf ACC (\%)\\
\midrule
ConSE~\cite{norouzi2013zero} & 44.5\\
Devise~\cite{frome2013devise} & 59.7\\
SYNC~\cite{changpinyo2016synthesized} & 46.6\\
SE-GZSL~\cite{Verma_2018_CVPR} & 69.2 \\
Gaussian-Ort~\cite{Zhang_2018_CVPR} & 70.5\\
GCNZ~\cite{wang2018zero} & 70.7 \\
\midrule
DGP (ours) & {\bf 77.3} \\
\bottomrule
\end{tabular}
\end{table}

\subsection{Comparison to Graph Attention Networks}

Table~\ref{tab:results2} illustrates the results for a 1-hidden-layer and 2-hidden-layer GCN with the attention mechanism proposed in GAT~\cite{velickovic2017graph}. Note, performance degrades compared to a 1-hidden-layer GCN (i.e. SGCN(-f)). The hidden dimension is 2048 and training settings are the same as in the paper.

\begin{table}[!h]
\def\arraystretch{0.6}
\small
    %\caption{Top-k accuracy for the different models on the ImageNet dataset.}
      \caption[t]{Accuracy on ImageNet for a 1- and 2-hidden-layer GAT~\cite{velickovic2017graph} compared to a 1-hidden-layer GCN without attention.}
      \label{tab:results2}
\begin{tabular}{l|c|ccccc}%c}
        %\cmidrule[1.5pt]{1-3}
        \toprule
        \bf \multirow{2}{*}{Test set} & \bf\multirow{2}{*}{Model} & \multicolumn{5}{c}{\bf Hit@k (\%)}\\
        & & 1 & 2 & 5 & 10 & 20\\
        \midrule
        %{\multirow{4}{*}{\bf 2-hops}} & GCNZ$^\ddagger$ & 19.8 & 33.3 & 53.2 & 65.4 & 74.6\\
        {\multirow{3}{*}{\bf 2-hops}} & GAT-1 & 24.1 & 37.5 & 57.2 &69.7 & 79.4\\
        & GAT-2 & 23.3 & 36.9 & 56.8 &68.7 & 77.9\\
        %& baseline & 23.45 & 36.78 & 56.43 & 68.29 & 77.29\\
        \cline{2-7}
        %\Tstrut & 1-layer GCN(-f) & 24.8 & 38.3 & 57.5 & 69.9 & 79.6\\
        \Tstrut & GCN-1 (ours) & {\bf 24.8} & {\bf 38.3} & {\bf 57.5} & {\bf 69.9} & {\bf 79.6}\\
        %& \ADGN(-wf) & 23.8 & 36.9 & 56.2 & 69.1 & 78.6\\
        %& \ADGN(-f) & 24.6 & 37.8 & 56.9 & 69.6 & 79.3\\
        %\Tstrut & SGCN (ours) & 26.2 & 40.4 & 60.2 & 71.9 & 81.0\\
        %& \ADGN(-w) & 25.4 & 39.5 & 59.9 & 72.0 & 80.9\\
        %& {\ADGN} (ours)& {\bf 26.6} & {\bf 40.7} & {\bf 60.3} & {\bf 72.3} & {\bf 81.3}\\
        \bottomrule
        \end{tabular}
\end{table}

\end{document}